# Real-Time Face & Eye Tracking and Blink Detection using Event Cameras

Cian Ryan[1*], Brian O'Sullivan[1], Amr Elrasad[1], Joe Lemley[2], Paul Kielty[2], Christoph Posch[3] and Etienne Perot[3]

[1] Xperi Corporation
[2] National University of Ireland Galway
[3] Prophesee
* Corresponding author: cian.ryan@xperi.com

## Abstract

Event cameras contain emerging, neuromorphic vision sensors that capture local - light intensity changes at each pixel, generating a stream of asynchronous events. This way of acquiring visual information constitutes a departure from traditional frame-based cameras and offers several significant advantages – low energy consumption, high temporal resolution, high dynamic range and low latency. Driver monitoring systems (DMS) are in-cabin safety systems designed to sense and understand a drivers physical and cognitive state. Event cameras are particularly suited to DMS due to their inherent advantages. This paper proposes a novel method to simultaneously detect and track faces and eyes for driver monitoring. A unique, fully convolutional recurrent neural network architecture is presented. To train this network, a synthetic event-based dataset is simulated with accurate bounding box annotations, called Neuromorphic-HELEN. Additionally, a method to detect and analyse drivers' eye blinks is proposed, exploiting the high temporal resolution of event cameras. Behaviour of blinking provides greater insights into a driver level of fatigue or drowsiness. We show that blinks have a unique temporal signature that can be better captured by event cameras.

**KEYWORDS**: Event Cameras, Convolutional Neural Network, Driver Monitoring System

## 1. Introduction

Event sensors or Dynamic vision sensors (DVS) are novel, bio-inspired asynchronous vision acquisition devices. In contrast to standard frame-based sensing (Mead, 1990), individual pixels asynchronously report "event" streams of intensity changes. Event cameras offer several significant advantages over conventional cameras including i) high temporal resolution (order of milliseconds), ii) high dynamic range (140 dB against standard camera 60 dB) and iii) low power consumption (Gallego et al., 2019; Mahowald, 1994). This paper leverages some of these advantages for applications in driver monitoring systems (DMS). Specifically, we propose novel methods to 1) detect and track facial features from events and 2) detect and analyse blinking patterns.

Pixels in an event camera asynchronously monitor light intensity and individually report when they detect a local relative intensity change, i.e. temporal contrast, exceeding a predefined threshold. This threshold crossing is called an "event". Event data encode the x, y coordinate of the reporting pixel, time of occurrence (time stamp) and polarity of the detected intensity change (+ or – for increase or decrease in intensity). This sensing paradigm implies that DVS pixels individually adapt their sampling rate to the scene dynamics, i.e. scene motion they observe. This is unlike conventional cameras that sample at a fixed rate. Event cameras are particularly suited to DMS due to their high temporal resolution, enabling some advanced DMS functionalities beyond the capabilities of standard cameras. This includes low latency eye tracking, blink analysis, faster object detection and potentially even crash assessment. Typically, DMS utilise standard RGB or near-infrared (NIR) cameras. However, exploring alternative modalities, including event camera DMS (E-DMS) solutions, is both significant and timely given that such systems are expected to become mandatory across Europe by 2022. Moreover, as we progress to higher levels of vehicle automation, understanding the drivers cognitive and physical capabilities will be increasingly more important as control alternates between human and machine (Ryan et al., 2019, 2020; Sheehan et al., 2017). Currently, there is a growing interest in the applicability of event cameras within the scope of autonomous vehicles (Binas et al., 2017; Chen, Cao, et al., 2020; Jianing Li et al., 2019; Maqueda et al., 2018; Nitti et al., 2020). However, limited attention has been given to event-based DMS.

Event cameras naturally respond to edges in a scene which can simplify the detection of lower level features such as key points and corners (Gallego et al., 2019). Exploiting this, we present a novel methodology to detect and track faces and eyes using event cameras. We propose a novel, fully convolutional recurrent YOLO architecture for multi-object detection and tracking, specifically to detect and track faces and eyes for DMS. The architecture is based on YOLOv3-tiny (You Only Look Once) (Redmon & Farhadi, 2018) and integrates a fully convolutional gated recurrent unit (GRU) layer. The vast majority of object detection literature focuses on single image detection. However, the natural sparsity of events demands a temporal connection. We will refer to the proposed architecture as a gated recurrent YOLO (GR-YOLO). We quantitatively evaluate the network performance on a manually annotated dataset taken from a Prophesee event camera with resolution, 720x1280 pixels. We also qualitatively test both the detection network and blink detection algorithm on natural driving data gathered from 5 subjects with the event camera mounted on the front windscreen.

In addition to face and eye tracking, a method to detect and analyse blinking patterns is presented and draws on the high temporal resolution of event cameras. Blinking is indicative of various human behaviours and states including drowsiness. This is particularly critical to DMS where driver attention monitoring is a key functionality. Analysing the details of blinking patterns, other than blink frequency, is difficult for most conventional cameras. At 30 frames per second (fps), a blink will typically last 5 frames and a frame rate over 100fps is likely required for accurate blink analysis (Picot et al., 2009). In this paper, we detect blinks at a higher temporal resolution and further investigate the decomposition of a single blink. To detect blinks, we track the distribution of events within the detected eye regions every 5ms and automatically identify abnormal event spikes generated by blinks. A key advantage of event cameras is that we are not limited to a fixed frame rate. In this paper, we track face and eyes at an adaptive frame rate and detect and analyse blinks at an equivalent of 200 fps. Different tasks require different temporal resolutions. However, this is a luxury not supported by conventional cameras.

The limited availability of event-based data is one of the most significant barriers inhibiting the use of machine learning methods with event cameras. Recent developments in synthetic event simulations facilitate the generation of large-scale event-based datasets, unlocking virtually any conventional camera dataset (Delbruck et al., 2020; Gehrig et al., 2020; Rebecq et al., 2018). Using open-source event camera simulators, this paper proposes Neuromorphic-Helen (N-Helen), an event camera dataset with precise facial landmark annotations based on the Helen dataset (Le et al., 2012); this dataset consists of static images with facial landmark annotations. Thus, we first simulate a video dataset from Helen images through random augmentations and transformations with 6 degrees of freedom (DOF). We then simulate synthetic event streams and simultaneously map facial landmarks to event-space to produce precise bounding box annotations for face and eyes. This methodology is not limited to the Helen dataset and can incorporate any labelled dataset, including other modalities such as near-infrared. This is the first attempt in literature to generate a large-scale dataset for event camera face and eye detection.

Event-based vision sensors record changes in pixel intensity, with a latency that depends on the absolute light level; naturally responding to motion and suppressing static parts of the scene. This poses a significant challenge for event-based object detection where slow moving objects generate sparse events and thus, offer less information. Detecting and tracking static or slow-moving objects is difficult but necessary for DMS applications. Often, most motion in the scene can be attributed to movement outside the car and not in the drivers face (i.e. the region of interest). We show that the proposed GR-YOLO can preserve information over a longer time period. The network maintains information from previously detected locations if little or no extra information is made available. Tracking algorithms such as Kalman filters (Kalman, 1960; Wojke et al., 2017) can also support more accurate location estimates. Moreover, event representations such as leaky time surfaces preserve spatial information over longer periods of time. Such methods can be used to supplement GR-YOLO predictions to tackle the problem of limited object motion.

This paper offers several key novel contributions:

1. A synthetic event camera dataset is generated with facial landmark annotations using existing RGB datasets. This is the first approach to generate a face and eye detection event-based dataset. We also show that training on these datasets generalizes well to real world examples.
2. A novel event-based GR-YOLO network is proposed to detect and track faces and eyes. This is the first neural network approach to detect face and eyes with event cameras.
3. We propose a novel method to detect blinks and analyse blinking patterns. This method facilitates the extraction of drowsiness related features (i.e. blink duration etc.) and utilises the higher temporal resolution offered by event cameras.
4. We propose a DMS that performs different tasks at different and adaptive frame rates, exploiting the asynchronous nature if event cameras. This is a feature not offered by conventional cameras and could be a considerable advantage to DMS. In a conventional DMS setup, face tracking and blink analysis will be performed at the same rate. However, the speed of motion between blinks and head movement differs fundamentally and should be treated accordingly. In our approach, blinks are detected and analysed at an equivalent of 200 fps whereas face and eye detection is performed at an adaptive rate dependent on scene dynamics.

The remainder of this paper is structured as follows: Section 2 explores recent literature surrounding object detection and blink analysis for event and conventional cameras. Sections 3 and 4 describe the synthetic dataset generation and proposed methodology to detect face, eyes and blinks. Sections 5 and 6 details the results of our experiments, demonstrating the efficacy of the proposed methodologies, and presents concluding remarks.

## 2. Related Work

### 2.1 Object Detection and Tracking

Object detection and object tracking is a major field of study in computer vision but is limited in the domain of event cameras. In particular, there are few deep learning-based approaches, likely due to the absence of large event-based datasets. This is especially true for face and eye detection.

Lenz et al. (Lenz et al., 2018) propose an event-based face detection and tracking algorithm using hand-crafted features to detect blinks and subsequently track face and eyes. They exploit the unique temporal signature of blinks in event space and track faces accordingly. Jiang et al (Jiang et al., 2020) propose a single-image YOLO object detector and a correlation filter tracker in event space. Kalman filters are used to fuse the results from both detector and tracker (Jiang et al., 2020). Cannici et al. (Cannici et al., 2019a) propose YOLE (You Only Look at Events), a YOLO object detection network for event-based object detection, using a leaky surface event representation as input. Li et al (Jia Li et al., 2017) present a method to extract motion invariant features using recursive adaptive temporal pooling and subsequently detect object using a R-CNN for hand recognition. This method addresses the problem of object detection given slower object motion speed. Feature maps from previous timesteps are pooled based on a learned weighting. Other studies detect moving objects as regions of interest, as the event camera naturally responds to motion, and subsequently classify (Cannici et al., 2019b; Ghosh et al., 2014). However, these methods ignore the problem of object detection at different motion speeds. Joubert D et al (Joubert et al., 2019) exploits proposes object detection at multiple frame rates. They integrate events over several time windows to detect fast- and slow-moving objects. Few research studies refer to this inherent advantage of DVS to facilitate multiple frame rates.

Other methods estimate an intensity image from events and subsequently apply existing object detection or object classification algorithms to these reconstructed images (Rebecq et al., 2019; Scheerlinck et al., 2020). Such approaches bridge the gap between standard cameras and event cameras. Iacono et al (Iacono et al., 2018) study the performance of off-the-shelf DL-based object detection algorithms on event representations to investigate whether event representations contain enough information to discriminate between objects. They use a hybrid frame-based and event camera setup, using the same lens. The authors used a single shot detector (SSD), pretrained on the COCO dataset (Lin et al., 2014). However, these methods are unsuitable for DMS as most events occur outside the region of interest (ROI) i.e. car background rather than facial features. Moreover, these methods require the additional step of reconstruction which is computationally expensive and can diminish the inherent advantages of event cameras.

With conventional frame-based cameras, most object detection research relates to single image detection, processing frames independently. However, the natural input for most real-world applications are video sequences. Video detection methods have become more popular in recent years and can be generally categorised into feature-level and box-level approaches. Box-level tracking methods are often referred to as tracking-by-detection. Detections are performed on single images and temporal associations are based on detection outputs (Han et al., 2016; Lu et al., 2017; Ning et al., 2017). In contrast, feature-level methods integrate image features (Broad et al., 2018; Liu & Zhu, 2018; Zhu et al., 2017). The proposed GR-YOLO adopts this feature-level approach.

**2.2 Blink Detection and Pattern Analysis**

Blink detection and blink analysis is well researched with the use of conventional cameras (Muller, 2019). The relationship between blinking patterns and driver drowsiness is also well established (Baccour et al., 2019). In particular, oculomotor parameters including blink duration, re-opening delay, blink interval and closure speed are proven to be significant factors in drowsiness estimation (Schleicher et al., 2008). That said, a frame rate of over 100 fps is needed for precise parameter measurements. This is beyond the capabilities of current DMS systems that typically operate at 30-60 fps. To detect blinks, most studies locate landmarks around the eyes and subsequently detect blinks based on the distance between upper and lower eyelids (Soukupova & Cech, 2016). However, these methods require accurate facial landmark estimation and a high framerate camera.

To date, limited attention has been given to the potential of neuromorphic event cameras for blink detection and blink analysis. Recently, Lenz et al (Lenz et al., 2018) detect blinks over the full image based on the characteristic distribution of events exhibited during a blink. Detections are based on sparse cross-correlation between the observed distribution of events and their blink model. However, the authors process the entire image and thus cannot accurately detect blinks if there is significant motion elsewhere in the field of view. Chen et al (Chen, Hong, et al., 2020) propose a method to detect blinks from event cameras for drowsiness estimation. Detections are performed through a two-stage filtering process to remove events unrelated to eye and mouth regions and detect blinks based on the spikes in event recordings. No eye tracking is used, and a clustering algorithm is employed to determine whether the spike in events is due to eyes or mouth. Anastasios et al (Angelopoulos et al., 2020) propose a hybrid frame and event-based eye tracking system. Based on a two-dimensional parametric eye model, the authors identify blinks based on the premise that blinks will deform the fitted eye ellipse.

3. Dataset

There are no available datasets for face and eye detection for event cameras. To overcome this, we generated a large synthetic event-based dataset using Helen (Le et al., 2012) and mapping facial landmark annotations to event space. We call this dataset Neuromorphic Helen (N-Helen). Our approach takes any existing RGB datasets comprising of single images with facial landmark annotations. First, a set of random transformations and augmentations are applied to each image, simulating a video sequence with homographic camera motion with 6-DOF. At the same time, using the same transformation matrix for each image, we transform the annotated facial landmarks. From this, face and eye bounding boxes can easily be obtained. This results in a video sequence with annotations for each frame. This is a novel approach enabling us to train CNNs to operate in event-space without intermediary intensity representations. We show later that this method enables network to generalize well to real event cameras.

To simulate events, we use ESIM, an open source event simulator (Rebecq et al., 2018). Gehrig et al. (Gehrig et al., 2020) also utilises ESIM but first leverage DL-based frame interpolation to adaptively up-sample low framerate videos. We also adopt this approach. Other simulators include V2E which models noise characteristics more accurately (Delbruck et al., 2020). However, for the purpose of this paper, we adopted ESIM. In order to simulate realistic events, we must carefully choose the contrast threshold (CT) and refractory period. Positive and negative CTs are sampled from $\sim N(0.2, 0.05)$. Refractory period is set to 1ms. Next, to match this continuous stream of synthetic events to frame-based annotations, we consider a window of events leading up to the frame timestamp. This window is based on a fixed duration interval. Figure 1 below illustrates the simulation of events and annotation mapping.

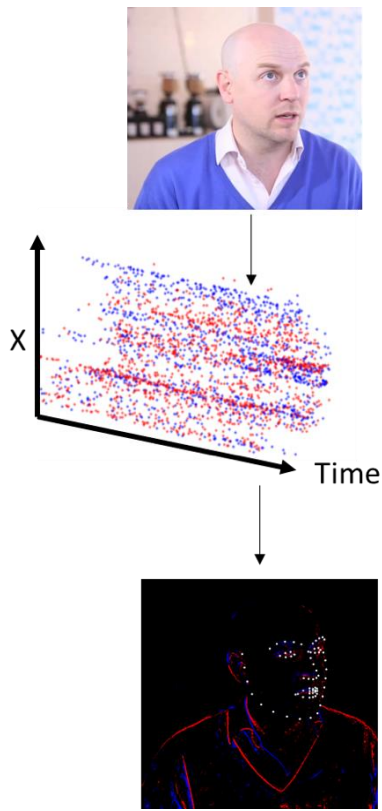

Figure 1. Illustration of RGB image to events and events to accumulated event representation with precise facial landmark annotations.

For the purpose of this paper, we generated a single static N-Helen dataset. Alternatively, events can be simulated on-the-fly. In particular, we can vary the contrast threshold, motion parameters etc. for each input before passing through the network. This facilitates an advanced augmentation technique and provides the network with unseen inputs at every epoch. A single image can generate a considerable amount of unique inputs.

## 4. Methodology

### 4.1 Event Representation

To process events using CNNs, events must be transformed into fixed size representation accepted by CNN. First, an event window needs to be established. Events within this window can be accumulated into a 2D frame. Typically, either a constant time window or adaptive time window with fixed number of events are considered. We adopt a fixed size event window with 50,000 events in each window at inference time for face and eye detection.

There are a number of different methods to create a frame-based event representation. One common event representation is voxel grids that encode both spatial and temporal information. With voxel grids, a fixed window size is discretized into $B$ temporal bins. With voxel grid representations, each event within this window shares its polarity between bins in a voxel as follows:

$$E(x_l, y_m, t_n) = \sum p_i \max(0, 1 - |t_n - t_i^*|)$$

Where $t_i^* = \frac{B-1}{\Delta T}(t_n - t_0)$ represents the normalized timestamp. In this paper, for training, we use a fixed event duration window in order to match event representations to ground truth labels defined at fixed time intervals. We tested B=5 and B=1, essentially reducing the voxel representation to simple event accumulation into a single channel. We found negligible performance difference using B=1 and with less computational cost. We also clipped the events between -10 and +10 to remove outliers. These outliers greatly affected performance and can be large. Values beyond this are aberrations.

There are several alternative representations that may improve performance further. For example, there are different kernels to equation 1 above. These kernels can even be learned in an end-to-end manner. Another representation is leaky time surfaces where events slowly decay over time.

### 4.2 Network Architecture

In this section, we describe the proposed fully convolutional gated recurrent YOLO network, outlined in Table 1 below. Convolutional and max pool layers (layers 0-11 in table 1.) down-sample the image resolution by a factor of $2^5$ = 32, resulting in a feature maps of size 8x8 during training. Each cell corresponds to an 8x8 section of the input image. Route layers indicate skip connections with feature map concatenation. GRU refers to the fully convolutional gated recurrent unit. Preceding the YOLO detection layers, 1x1 convolutions are used. The shape of the kernels is $1 \, x \, 1 \, x \, (B \, x \, (5 + C))$ where *B* is the number of predicted bounding boxes and *C* is the number of classes. B is set to 3 and C is 2 (face + eye). That is, the network predicts 3 bounding boxes at each cell. YOLO detection layers (layers 17 and 24) makes predictions on each cell for each box. Each YOLO detection layer predicts box coordinates and dimensions, objectness and class probabilities. Objectness and class probabilities reflect the probability that an object is contained within a bounding box and the conditional probability of a class given an object, respectively.

| LAYER | TYPE | FILTER | KERNEL/STRIDE | INPUT | OUTPUT |
|---|---|---|---|---|---|
| 0 | Conv | 16 | 3 / 1 | 256 x 256 x 1 | 256 x 256 x 16 |
| 1 | Maxpool | | 2 / 2 | 256 x 256 x 16 | 128 x 128 x16 |
| 2 | Conv | 32 | 3 / 1 | 128 x 128 x 16 | 128 x 128 x 32 |
| 3 | Maxpool | | 2 /2 | 128 x 128 x 32 | 64 x 64 x 32 |
| 4 | Conv | 64 | 3 /1 | 64 x 64 x 32 | 64 x 64 x 64 |
| 5 | Maxpool | | 2 / 2 | 64 x 64 x 64 | 32 x 32 x 64 |
| 6 | Conv | 128 | 3 / 1 | 32 x 32 x 64 | 32 x 32 x 128 |
| 7 | Maxpool | | 2 / 2 | 32 x 32 x 128 | 16 x 16 x 128 |
| 8 | Conv | 256 | 3 / 1 | 16 x 16 x 128 | 16 x 16 x 256 |
| 9 | Maxpool | | 2 / 2 | 16 x 16 x 256 | 8 x 8 x 256 |
| 10 | Conv | 512 | 3 / 1 | 8 x 8 x 256 | 8 x 8 x 512 |
| 11 | Maxpool | | 2 / 1 | 8 x 8 x 512 | 8 x 8 x 512 |
| 12 | Conv | 1024 | 3 / 1 | 8 x 8 x 512 | 8 x 8 x 1024 |
| 13 | Conv | 256 | 1 / 1 | 8 x 8 x 1024 | 8 x 8 x 256 |
| 14 | GRU | 256 | 3 / 1 | 8 x 8 x 256 | 8 x 8 x 256 |
| 15 | Conv | 512 | 3 / 1 | 8 x 8 x 256 | 8 x 8 x 512 |
| 16 | Conv | 21 | 1 / 1 | 8 x 8 x 512 | 8 x 8 x 21 |
| 17 | **YOLO** | | | 8 x 8 x 21 | **192 x 7** |
| 18 | **Route 14** | | | | 8 x 8 x 256 |
| 19 | Conv | 128 | 1 / 1 | 8 x 8 x 256 | 8 x 8 x 128 |
| 20 | Up-Sampling | | | 8 x 8 x 128 | 16 x 16 x 128 |
| 21 | **Route 20 8** | | | | 16 x 16 x 384 |
| 22 | Conv | 256 | 3 / 1 | 16 x 16 x 384 | 16 x 16 x 256 |
| 23 | Conv | 21 | 1 / 1 | 16 x 16 x 256 | 16 x 16 x 21 |
| 24 | **YOLO** | | | 16x16x21 | **768 x 7** |

Table 1. GR-YOLO network architecture

Layers 0 to 13 consist of a series of convolutional and max pooling layers. These layers encode information from the current reference frame and map to feature space. These layers process information from the reference frame only. The GRU is positioned after these layers to supplement information gained from the reference frame. As a result, the GRU offers higher-level feature relating to pertinent face and eye characteristics. The combined information is then propagated through a single convolutional layer before the first YOLO detection layer. To minimize complexity, a single GRU was employed. The benefits of additional recurrent layers at varying positions is subject to future work, but beyond the scope of this paper.

To make bounding box predictions, the YOLO layers employ anchor boxes, a set of predefined bounding boxes with set heights and widths. Anchors are essentially bounding box priors. They are configured to capture the scale and aspect ratio of the object classes relating to the current dataset and task at hand. As mentioned, at each cell, 3 anchor boxes are used. So, within a cell, a prediction is made for each anchor, based on the 1x1 convolutions explained above. The output is $(t_x, t_y, t_w, t_h, t_o, p_1, p_2)xB$ for each grid cell, where $t_o$ reflects objectness i.e. the probability a box contains an object. $p_1$ and $p_2$ represents the probability of each classes. Input to YOLO detection layers comprise of 21 feature maps: 7 x 3 = 21. The 3 relates to the 3 anchor boxes. The 7 relates to x, y centre coordinates, height, width, objectness and 2 class probabilities predicted from the previous 1x1 convolution. The equations below describe how this output is transformed to bounding box predictions:

$$b_x = \sigma(t_x) + c_x$$

$$b_y = \sigma(t_y) + c_y$$

$$b_w = p_w e^{t_w}$$

$$b_h = p_h e^{t_h}$$

Where $(b_x, b_y, b_w, b_h)$ represent bounding box centre x, y coordinates, width and height, $\sigma$ signifies the sigmoid function, $p_w$ and $p_h$ are bounding box prior width and height and $c_x$ and $c_y$ are the coordinates of the top left corner of the grid cell. Rather than predict absolute width and height, the model predicts width and height ($t_w$ and $t_h$) as log transforms or offsets to these predefined anchors. Offsets are applied to anchors boxes to create new width and height predictions. Essentially, dimensions are predicted by applying log-space transformations and subsequently multiplying by anchors. Centre coordinate predictions ($t_x$ and $t_y$) represent offsets relative to the top left corner of each cell ($c_x$ and $c_y$). The centre coordinates are transformed using a sigmoid function to force the output between 0-1 and within the cell. Objectness predictions ($t_o$) are also passed through a sigmoid function and interpreted as a probability. Figure 2 illustrates this process.

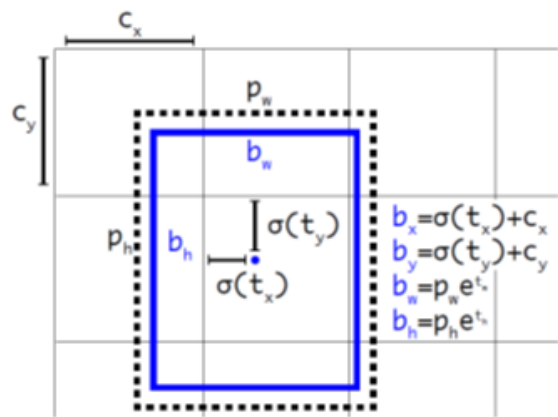

Figure 2. Bounding box with dimension priors and predictions (Redmon & Farhadi, 2018)

The network makes predictions at two different scales i.e. 8x8 and 16x16. At each scale, the network makes predictions for each the 3 anchors over each cell. This amounts to 6 anchors used over the 2 scales. Multiscale detection supports the detection of small and larger objects.

The network produces ((8 x 8) + (16 x 16)) x 3 = 960 predictions. Typically, there is only 1 face and 2 eyes in the field of view and so further filtering of the predictions is necessary. Filtering is first performed based on objectness scores. Boxes with low objectness probabilities (i.e. < 0.6) are removed. Objectness refers to the likelihood the bounding box contains an object. Non-maximum suppression is then used to further filter overlapping detections of the same object.

The fully convolutional GRU is positioned after the series of convolutional and max pooling layers (layer 14) just before the detection layers. One more convolutional layer IS used to further process the cumulative information. The GRU inputs and outputs 256 feature maps. The equations governing fully convolutional GRUs are as follows:

$$z_t = \sigma(W_z * x_t + U_z * h_{t-1})$$
$$r_t = \sigma(W_r * x_t + U_r * h_{t-1})$$
$$\tilde{h}_t = tanh(W * x_t + U * (r \odot h_{t-1}))$$
$$h_t = (1 - z_t)h_{t-1} + z_t\tilde{h}_t$$

where * is the convolution operator, $\odot$ is the Hadamard product. $x_t$ is the input at time t, $z_t$ is the update gate, $r_t$ is the reset gate, $\tilde{h}_t$ is the candidate activation, $h_t$ is the output, $\sigma$ is the sigmoid function and $W_z, U_z, W_r, U_r, W$ and $U$ are the learnable weights.

During inference, an input size of 288x512 is used, reflecting the resolution of the event camera (720x1280) divided by 2.5 in both directions. The fully convolutional network can take any input size if the width and height are divisible by $2^5$. The network consists of approximately 12.8 million parameters. The original YOLOv3-tiny contains 5.56 billion flops (GFLOPS) and the additional GRU cell contains 0.26 GMACS (roughly 0.13 GFLOPS).

### 4.3 Training Details

A total of 2330 video sequences were used during training. The dataset was split randomly into 2000 training sequences and 330 validation sequences. The network was implemented with PyTorch. An ADAM optimizer was used with a learning rate of 1x10$^{-3}$ reduced by a factor of 0.8 every 20 epochs for a total of 120 epochs. Weight decay of 1x10$^{-3}$ was applied. Mean squared error (MSE) loss was used and calculated over both YOLO detection layers.

Random data augmentations were also applied. A key problem in event-based object detection is reliance on object motion. To address this, we augment the training input sequence by randomly feeding sequences of zeros into the network to reflect no motion in the face region. This encourages the network to weight previous predictions more when the current input lacks information. The network is forced to remember where the face and eyes are based on previous predictions. Data augmentations such as random rotations and horizontal flipping were applied to foster rotational and geometric invariance.

### 4.4 Blink Detection

The natural polarity output of event cameras is particularly suited to the detection of rapid movements such as blinks. Positive and negative polarities indicate an increase or decrease in pixel intensity above a predefined threshold. Blinks generate a significant number of events within the eye regions. In particular, the downward movement of the eyelid over the eye typically generates an abnormal surge in the number of positive and negative events. Figure 3 below shows a blink with blue and red indicating positive and negative polarities respectively. Specifically, it shows the upward motion of the eye lid at the end of a blink. The polarities are reversed during the eyelid downward movement.

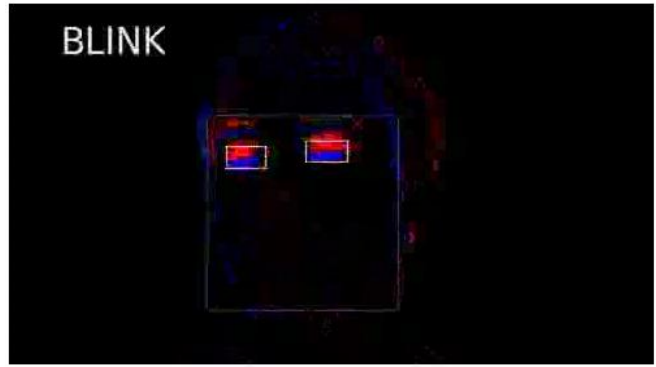

Fig 3. A typical blink presented in terms of +/- polarities

The proposed process of detecting and analysis blinks exploits the characteristic and salient features of a blink and is as follows:

1. Using a fixed time window of 5ms, we identify points in time where the mean polarity per pixel in either eye region is above a predefined threshold. This threshold is a function of the object distance from the camera. Naturally, the closer the object is to the camera, the more events generated. This first step is designed to filter noise and events due to head motion. See Figure 4, top right time series. The positive (top) and negative (bottom) sides of the time series represent the average polarity per pixel i.e. + and -. Orange and blue colours represent right and left eyes respectively. A bimodal distribution is evident and characteristic of a typical blink.
2. Next, we filter identified points in time from step 1 based on the distribution of polarity along the vertical axes for either eye crop region. Specifically, we look at the standard deviation of the distribution of events from the perspective of the vertical axis. The standard deviation is shown in a time series in Figure 4, bottom right. Large standard deviation indicates horizontal motion as the positive and negative polarities are separated horizontally, see Figure 4. This removes horizontal pupil movements that may have been identified as blinks. Fast pupil movement or saccades generate a significant number of events and so it is important to distinguish between horizontal blink movements and other movements.
3. Lastly, using the same fixed duration event window of 5ms, we can decompose the blink at a granular level. The distribution of events, as seen in the time series of Figure 4 right hand side, is typically bimodal. Each modal represents the downward and upward motion of the eyelid. Additional features can be extracted including blink duration, eyelid closing/opening duration, eyelid closed time and even speed of a blink.

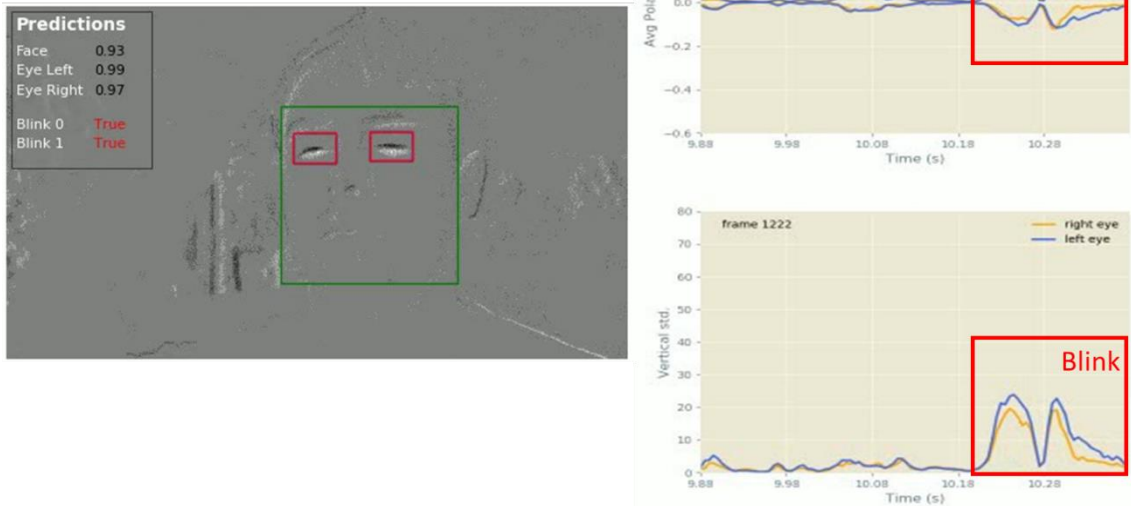

*Figure 4. An example of a blink (left) and both timeseries (right) representing different aspects of the distribution of events within each eye crop region. Top right is the mean polarity per pixel. Bottom right is the standard deviation along the vertical axis.*

## 5. Experiments and Results

Video results of the proposed face and eye tracking and blink detection are included in the supplementary material. For data privacy reasons, we show just one subject video. This video is slowed down considerably to better visualize the predictions.

To test the efficacy of the proposed face and eye detector, a set of test videos were recorded and manually annotated. This supports a quantitative evaluation of the proposed GR-YOLO network. Annotations include bounding box coordinates for face and eyes and blink timestamps. The proposed methodology lies in a relatively unexplored area of research and so the absence of a diverse benchmark dataset limits significant testing. Lenz et al. (Lenz et al., 2018) compared face detection performance against standard frame-based algorithm applied to reconstructed intensity images. However, this is not a fair comparison as these algorithms were not designed to operate on these reconstructions and rely heavily on high quality reconstruction estimates.

Naturalistic driving data was also recorded to test the performance of both the face and eye tracking and blink detection. These video results can be found in the supplementary material. Eye blinks are annotated manually in these videos to evaluate the precision of our blink detection algorithm.

### 5.1 Face/Eye Detection and Tracking

Three test videos were recorded from a Prophesee event camera to test the performance of the GR-YOLO network. Bounding box labels were manually annotated for the face and eyes in each frame. Annotating is a time-consuming and labour-intensive process. In our experiments, a single test subject is used in order to obtain comparative performances across test videos. It is important to note that this test subject was not in the training data. Test videos include:

1. Subject exhibiting slight head movements – 189 equivalent frames

2. Subject exhibiting head movements with larger yaw and pitch angles – 448 equivalent frames
3. Subject exhibiting slight head movements with another fast-moving object within the field of view – 127 equivalent frames

Examples of each test video are shown in figure 5 below. Test videos 1 and 2 are to demonstrate the network ability to handle different head positions. As this is designed for DMS, we kept the subject at approximately 60cm (typical distance in a car setting with front facing camera). Test video 3 is used to test a scenario where most of that motion is dominated by a different and irrelevant object. Thus, the detail in the face region is minimal or non-existent. The rightmost image contains a moving object (hand and pen) but no face information, although the face is in the field of view. Test video 2 contains more equivalent frames due to the higher degree of motion exhibited by the subject.

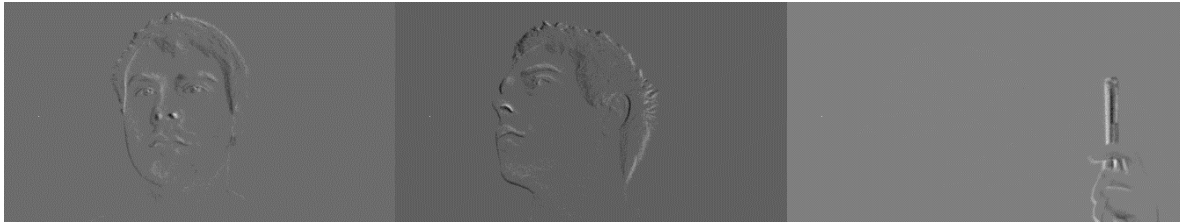

*Figure 5. Test videos 1, 2 and 3 shown from left to right, each ranging in difficulty.*

Quantitative results from the three test videos are shown in table 2 below. Naturally, the detection network performs comparatively worse with larger head movements. This is probably due to the nature of the training data. The Helen dataset consists mostly of front facing subjects with less examples of larger yaw and pitch head pose angles. However, this can certainly be improved by adding additional datasets with a wider range of angles to training. Moreover, we can see that the network continues to perform well for test 3 despite the lack of face information. This can be more easily visualized in Figure 6 below. The performance here can be attributed to our augmentation technique where zero inputs were fed through the network after some time. This encouraged the network to remember the last know position of the face given limited new information. Tracking with Kalman filters or other method would improve these results further but for an unbiased analysis of the network, we do not employ post-detection tracking.

|  | Test 1 | Test 2 | Test 3 |
|---:|:---:|:---:|:---:|
| *Average MSE Loss* | 4.42 | 7.15 | 4.45 |
| *Mean Average Precision (MAP)* | 0.95 | 0.86 | 0.89 |
| *Average Recall* | 0.95 | 0.86 | 0.89 |

*Table 2. Performance of GR-YOLO face and eye detection network based on average mean squared error, mean average precision and average recall.*

Example qualitative results are shown in figure 6 below for test videos 1 (top), 2 (middle) and 3 (bottom). Green bounding boxes represent ground truth. Face and eye detection are represented by purple and pink bounding boxes respectively.

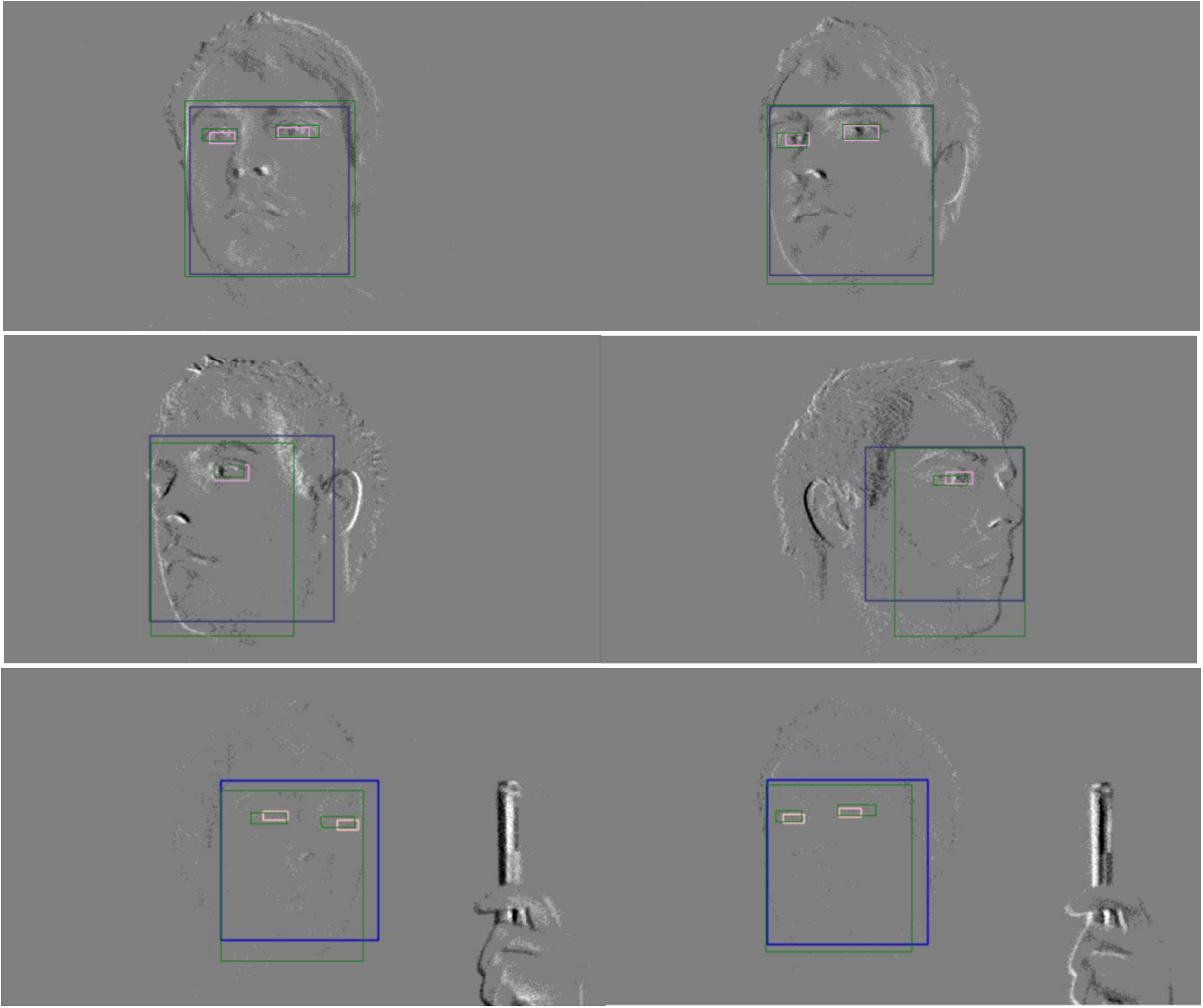

*Figure 6. Example outputs for test videos 1, 2 and 3 shown in rows 1, 2 and 3 respectively. Green boxes indicate ground truth.*

There are no available event camera datasets with annotated face bounding boxes. Our performance evaluation is constrained due to the data limitations. Future data acquisitions and annotations is required to perform more statistically significant testing. However, this evaluation still provides a preliminary quantitative evaluation and demonstrates the efficacy of our network.

3 subjects were recorded to test the performance of our GR-YOLO detector and blink detection algorithm in unconstrained driving environments. Each subject was recorded while driving. Recording was started at a random time during the trip and we continued to record until a sufficient number of blinks were observed. Eye blinks are manually annotated. The evaluation metrics used are precision and recall, formulated as:

$$Precision = \frac{True\ Positive}{True\ Positive + False\ Positive}$$

$$Recall = \frac{True\ Positive}{True\ Positive + False\ Negative}$$

Table 3 below illustrates the performance of our blink detection algorithm on our driving test datasets.

|  | Sub 1 | Sub 2 | Sub 3 |
|---:|:---:|:---:|:---:|
| *Time (seconds)* | 141 | 60 | 132 |
| *No. Blinks* | 43 | 43 | 59 |
| *True Positives* | 42 | 42 | 49 |
| *False Negatives* | 1 | 1 | 10 |
| *False Positives* | 7 | 1 | 12 |
| *Precision* | 85.7 % | 98 % | 83.1 % |
| *Recall* | 97.7 % | 98 % | 80.3 % |

*Table 3. Performance of blink detection algorithm based on precision and recall for 3 subjects*

Subject 3 exhibits relatively more false positives. Some of these are due to the head moving upward and downward as a result of bumps in the road. Vertical up and down head motion often generates similar event distributions as blinks. However, we could account for this by calculating head pose angles or the motion of the head. By knowing the movement of the head, we could remove these false detections. However, this is currently beyond the scope of this paper. Subject 2 exhibited abnormal blink frequencies, but this is likely due to the open window and incoming wind aggravating the eye and forcing excessive blinking. Example results from our event-DMS are shown in Figure 7 below.

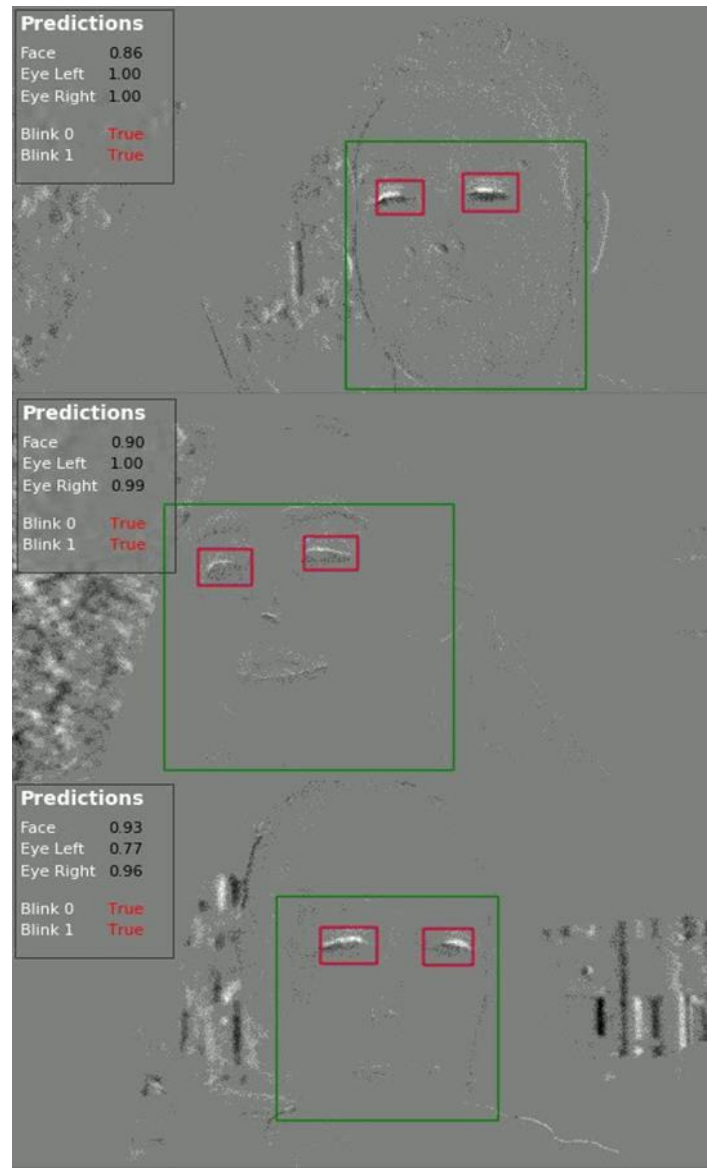

*Figure 7. Example detected blinks from driving scenario with mounted camera on windscreen for subject 1 (top), subject 2 (center) and subject 3 (bottom).*

Lenz et al. (Lenz et al., 2018) are the first to utilize the natural temporal signature of eye blinks using event cameras. The objective is to detect faces from detected blinks given that unique temporal signature of eye blinks are easily detected by event cameras. The authors detect blinks based on the correlation between local eye activity against a temporal model of blinks. First, an integrated frame is created with decaying events. The frame is then split into $n \times n$ tiles, where each tile is monitored to checked whether it correlates with a canonical blink model. The blink model is created based on 120 blinks over 6 subjects and over a window of 250ms. Sparse cross-correlation is used to determine a cross-correlation score and classify blinks. In contrast to this method, our approach first detects faces and eyes and uses cropped eye regions to identify and analyse blinking behaviours. The authors declare an average blink detection rate of approximately 55%.

Experimental results for blink analysis and oculomotor feature extraction are primarily qualitative. In Figure 8 below, we have isolated a single blink using a fixed duration window of 5ms. Using the detected time points from the blink detection methodology, we can model the bimodal distribution and extract oculomotor features such as duration, speed, closing time etc. However, the automatic extraction of such features is currently beyond the scope of this paper. However, we can see from the isolated blink in Figure 8 that the extraction of such features should be straightforward. Pupil movement and saccadic eye motion also introduce additional challenges that need to be addressed.

A typical blink exhibits this bimodal distribution for both positive and negative polarities. The initial downward eyelid movement results in a peak in positive and negative events. The bottom of the blink (eye closed), represents a point in the blink where the eye lid velocity reaches zero and no events are generated. Lastly, the upward turn causes another spike in positive and negative events. We found that the upward movement is typically longer than the eyelid closing. The high temporal resolution of event cameras allows us to record the entire blink and decompose it into individual components.

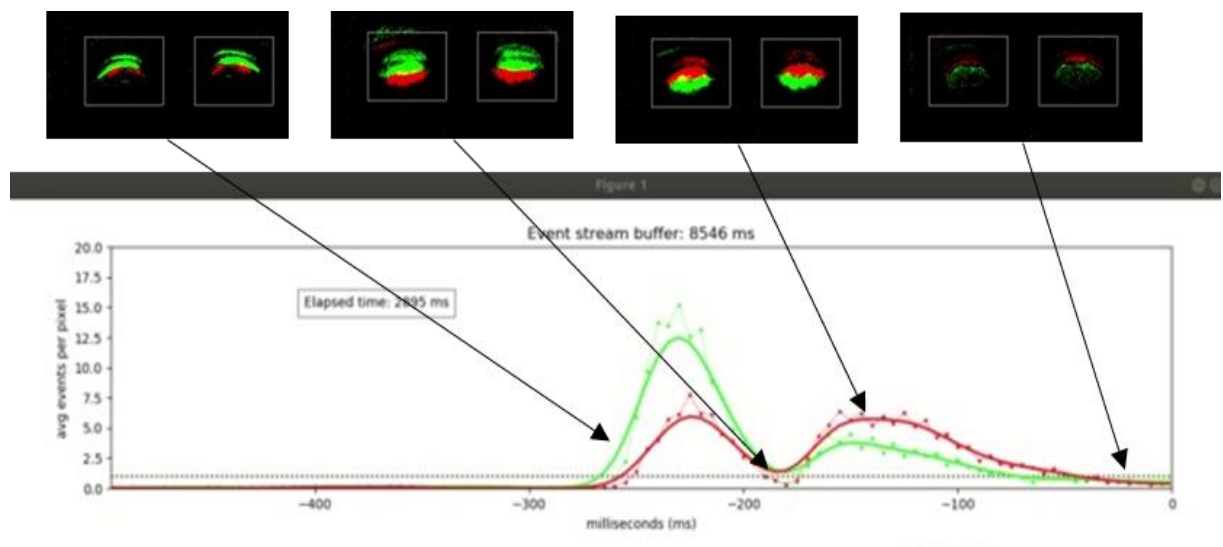

*Figure 8. Example blink decomposed into individual components.*

## 6. Conclusion

This paper presents a novel methodology to detect and track face and eyes for event cameras using a novel fully convolutional recurrent neural network. In addition, a lightweight method to detect blinks is presented. Performance of both methods are tested qualitatively and quantitatively on test datasets collected manually. Currently, there are no benchmark datasets to compare to. That said, the results expressed in this paper are significant and demonstrate the applicability of event cameras for DMS. Moreover, we demonstrated the efficacy of the presented methodology in unconstrained driving environments. Ultimately, event cameras offer several key advantages over conventional cameras making them particularly suitable for driver monitoring systems. In particular, the higher temporal resolution and the ability to dynamically adapt framerates per task. Moreover, their natural response to object motion affords a more direct approach to blink detection. These characteristics support an advanced DMS with features beyond the capabilities of current fixed framerate solutions. The capabilities of event camera-based driver monitoring are not only limited to the features proposed in this paper and other potential applications include low-latency eye gaze tracking, saccadic eye motion detection and collision analysis.

## Acknowledgments

This work is supported by Xperi Corporation and Prophesee. We would like to thank Dr. Petronel Bigioi and Dr. Chris Dainty of Xperi for their support throughout this project.